\documentclass[journal]{IEEEtran}

\usepackage{graphicx}
\usepackage{cite}
\usepackage{amsmath}
\usepackage{amssymb}
\usepackage{float}
\usepackage{multirow}
\usepackage{array}
\usepackage{booktabs}
\usepackage{dutchcal}
\begin{document}

\title{Group-Aware Graph Neural Network for Nationwide City Air Quality Forecasting}

\author{Ling Chen, Jiahui Xu, Binqing Wu, Yuntao Qian, Zhenhong Du, Yansheng Li, Yongjun Zhang

\thanks{This work was supported by the National Key Research and Development Program of China under Grant 2018YFB0505000. (Corresponding author: Ling Chen.)}

\thanks{Ling Chen, Jiahui Xu, and Binqing Wu are with the College of Computer Science and Technology, Alibaba-Zhejiang University Joint Research Institute of Frontier Technologies, Zhejiang University, Hangzhou 310027, China (emails: lingchen@cs.zju.edu.cn, xujiahui19@zju.edu.cn, binqingwu@zju.edu.cn).}

\thanks{Yuntao Qian is with the College of Computer Science and Technology, Zhejiang University, 38 Zheda Road, Hangzhou 310027, China (email: ytqian@zju.edu.cn).}

\thanks{Zhenhong Du is with the School of Earth Sciences, Zhejiang University, Hangzhou 310058, China (e-mail: duzhenhong@zju.edu.cn).}

\thanks{Yansheng Li and Yongjun Zhang are with the School of Remote Sensing and Information Engineering, Wuhan University, Wuhan 430079, China (e-mails: yansheng.li@whu.edu.cn,  zhangyj@whu.edu.cn).}

}

\markboth{$>$ REPLACE THIS LINE WITH YOUR PAPER IDENTIFICATION NUMBER (DOUBLE-CLICK HERE TO EDIT) $<$}%
{Shell \MakeLowercase{\textit{et al.}}: Bare Demo of IEEEtran.cls for IEEE Journals}

\maketitle

\begin{abstract}
The problem of air pollution threatens public health. Air quality forecasting can provide the air quality index hours or even days later, which can help the public to prevent air pollution in advance. Previous works focus on citywide air quality forecasting and cannot solve nationwide city forecasting problem, whose difficulties lie in capturing the latent dependencies between geographically distant but highly correlated cities. In this paper, we propose the group-aware graph neural network (GAGNN), a hierarchical model for nationwide city air quality forecasting. The model constructs a city graph and a city group graph to model the spatial and latent dependencies between cities, respectively. GAGNN introduces differentiable grouping network to discover the latent dependencies among cities and generate city groups. Based on the generated city groups, a group correlation encoding module is introduced to learn the correlations between them, which can effectively capture the dependencies between city groups. After the graph construction, GAGNN implements message passing mechanism to model the dependencies between cities and city groups. The evaluation experiments on Chinese city air quality dataset indicate that our GAGNN outperforms existing forecasting models.
\end{abstract}


\IEEEpeerreviewmaketitle

\section{Introduction}

\IEEEPARstart{A}{ir} pollution causes a range of health problems, particularly harming the cardiopulmonary system. The air quality index (AQI) provides a quantitative description of air quality conditions and is calculated from the concentration of air pollutants in an area. Forecasting the AQI values of nationwide cities can help the government combat air pollution and the public perceive local air quality trends in advance, especially during major air pollution events, e.g., dust storms and wildfires.

Nationwide city air quality forecasting, as a typical spatial-temporal forecasting problem, involves multiple challenges. First, AQI values have complex dependencies in temporal dimensions with implied periodicity and trendiness. Second, AQI values have complex dependencies in spatial dimensions with the dependencies between both geographically adjacent and distant cities.

Existing air quality forecasting models can be divided into two categories: physical models and machine learning models. Physical models \cite{p1,p22,p28} are designed based on the theories of air motion and matter diffusion, which require the integrity of pollution source data and have poor generalization ability. Machine learning models learn the relationship between input features and AQI values from the data and can be further subdivided into time series analysis models, statistical regression models, and deep learning models. Time series analysis models \cite{p11} forecast AQI values by finding linear patterns in the historical observation series, but cannot accept the feature inputs other than sequence data. Statistical regression models \cite{p26,p32} are capable of supporting multi-source feature inputs, but the performance of models depends on feature engineering. Deep learning models can automate feature learning by stacking multiple neural networks to fit the nonlinear transformation from inputs to outputs. These models generally employ recurrent neural networks (RNNs) \cite{p16,p20,p29} and their variants to model complex dependencies in temporal dimensions, and convolutional neural networks (CNNs) \cite{p2} or graph neural networks (GNNs) \cite{p6,p17,p20,p25,p29} to model complex dependencies in spatial dimensions. However, most of these models focus on citywide air quality forecasting and do not consider the dependencies between cities. A few works \cite{p20,p25,p29} model the spatial dependencies between cities, i.e., geographically adjacent cities have similar air quality, but ignore the latent dependencies between geographically distant but highly correlated cities, e.g., the air quality of coastal cities is affected by sea breezes \cite{p18}. The receptive field of cities in these works is limited, which is a common problem in spatial-temporal forecasting.

For those models that employ GNNs, a naive strategy that expands the receptive field of entities is to deepen GNNs. However, recently studies \cite{p12,p15,p23} discovered that GNNs suffer from the over-smoothing issue when going deeper, i.e., the representations of adjacent entities converge and their local features are lost.

Hierarchical graph neural networks (HGNNs) are a kind of GNNs that construct multi-level graphs \cite{p14,p27,p29,p31} and implement interactions among multi-level graphs to model the dependencies between entities. HGNNs expand the receptive field of entities by constructing multi-level graphs rather than deepening GNNs, thus relieving the conflict between expanding receptive fields and preserving local features. However, existing HGNNs rely on predefined rules to construct the coarsened graph, which cannot effectively capture the latent dependencies between entities.

In this paper, we propose the group-aware graph neural network (GAGNN), a hierarchical model for nationwide city air quality forecasting. The main contributions of our work are summarized as follows:

(1) We propose GAGNN, which constructs a city graph and a city group graph to model the spatial and latent dependencies between cities, respectively.

(2) We introduce the group correlation encoding module, which end-to-end learns the correlations between city groups to effectively capture the dependencies between city groups.

(3) We evaluate GAGNN on the Chinese city air quality dataset and compare it with the SOTA spatial-temporal forecasting models. The experimental results indicate that GAGNN outperforms existing models.

\section{Related Work}
\subsection{Air Quality Forecasting}
Existing air quality forecasting models can be divided into two categories: physical models and machine learning models. Physical models, e.g., street canyon model \cite{p22,p28} and Gaussian plume model \cite{p1}, are designed based on the theories of air motion and matter diffusion, which forecast air quality by simulating the emission and diffusion processes of air pollutants. However, these models require the integrity of pollution source data and have poor generalization ability.

Machine learning models learn the relationship between input features and AQI values from data and can be further subdivided into time series analysis models, statistical regression models, and deep learning models. Time series analysis models forecast AQI values by finding linear patterns in the historical observation series. Lee et al. \cite{p11} introduced ARIMA, a time-series analysis model, to forecast air quality. Time series analysis models cannot utilize other influencing factors, e.g., locations, as they cannot accept the feature inputs other than sequence data. Statistical regression models are capable of supporting multi-source feature inputs. Yu et al. \cite{p32} introduced the random forest model to capture the complex nonlinear relationships between multi-source influencing factors and air quality. Wang et al. \cite{p26} utilized RBF network and SVR to forecast air quality, further introducing PCA to reduce the dimensionality of input data. These models are simply used to adapt different feature inputs to different tasks, and their performances depend on feature engineering.

Deep learning models can automate feature learning by stacking multiple neural networks to fit the nonlinear transformation from inputs to outputs. Liang et al. \cite{p16} proposed GeoMAN, an encoder-decoder based spatial-temporal forecasting framework, which uses an attention mechanism to model the correlations between different metrics at the same sensor and the same metric at different sensors. Chen et al. \cite{p2} proposed PANDA, a multi-task air quality modeling framework, which implements air quality forecasting and air quality estimation in one model. Qi et al. \cite{p21} embedded feature selection and semi-supervised learning methods in deep neural networks, using unlabeled spatial-temporal information to improve the forecasting performance. Recently, GNNs have become practical tools for modeling non-Euclidean distributed entities. Qi et al. \cite{p20} proposed GC-LSTM, which introduces the graph convolution network (GCN) to capture the dependencies among air quality monitoring stations. Lin et al. \cite{p17} proposed GC-DCRNN, which constructs the graph based on the geographic context similarity between monitoring stations, and combines the diffusion convolution operation with the GRU gate. Ge et al. \cite{p6} constructed several graphs based on different similarity metrics between monitoring stations, modeling the correlations between stations in different semantic spaces. Wang et al. \cite{p25} introduced prior knowledge into the graph construction process, enabling the massage passing process in the graph perceiving weather factors. Xu et al. \cite{p29} proposed HighAir, which constructs a city graph and station graphs to consider the city-level and station-level patterns of air quality, respectively.

Most of the above models focus on citywide air quality forecasting and do not consider the dependencies between cities. A few works \cite{p20,p25,p29} model the spatial dependencies between cities, but ignore the latent dependencies between geographically distant but highly correlated cities.

\subsection{GNNs to Expand the Receptive Field}
Expanding the receptive field of entities enables GNNs to capture the dependencies between distant entities. To achieve this, a naive strategy is deepening GNNs \cite{p34}. However, message aggregation and representation updating in each GNN layer make the representations of adjacent entities more similar. The local features of entities would be lost when GNNs go deeper, and this issue is called over-smoothing \cite{p15}.

There are two types of methods to expand the receptive field without raising the over-smoothing issue: improving the architectural designs of GNNs \cite{p12,p13,p23} and introducing HGNNs. The former methods preserve the local features of nodes by introducing some specific designs of GNNs. Rong et al. \cite{p23} randomly removed a certain number of edges at each training epoch to alleviate both over-smoothing and over-fitting issues. Li et al. \cite{p12} referred to ResNet \cite{p7} and introduced skip connections in GNNs. Built on the previous work, Li et al. \cite{p13} further introduced message normalization and proposed a pre-activation version GNN. The receptive field size, i.e., the layer number of GNNs, of these methods is fixed, which is not adaptive for each entity in the graph.

HGNNs construct multi-level graphs by graph pooling methods \cite{p4,p10,p30} to relieve the conflict between expanding receptive fields and preserving local features. Yu et al. \cite{p31} proposed ST-UNet, which introduces a heuristic graph pooling method to construct the coarsened graph. Li et al. \cite{p14} utilized a geography-based HGNN to model a geographic information system, learning the relationship between socio-economic Census data and election results. Wu et al. \cite{p27} proposed HRNR, which constructs multi-level graphs to learn the representations of road segments. Zhang et al. \cite{p33} proposed SHARE, which combines HGNNs with semi-supervised learning to forecast citywide parking availability. However, existing models rely on predefined rules to construct the coarsened graph, i.e., using the geographic distance between nodes in the basic graph to define the correlations between nodes in the coarsened graph, which cannot effectively capture the latent dependencies between entities.

\section{Methodology}

\subsection{Definitions}
\textbf{Cities and city groups:} We define $C=\{c_i\}_{i=1}^{N_{\text{city}}}$ as the set of cities, $\mathbf{L}\in \mathbb{R}^{N_{\text{city}} \times 2}$ as the location matrix of cities, i.e., longitude and latitude, and $G=\{g_i\}_{i=1}^{N_{\text{group}}}$ as the set of city groups, where $N_{\text{city}}$ denotes the number of cities and $N_{\text{group}}$ denotes the number of city groups.

\textbf{City graph and city group graph:} We define $\mathcal{g}=(V, A, \mathbf{X}, \mathbf{E})$ as the city graph, where $V$ denotes the set of city nodes, $A$ denotes the set of edges, $\mathbf{X}$ denotes the node attribute matrix, and $\mathbf{E}$ denotes the edge attribute matrix. We define $\mathcal{G}=(\mathcal{V},\mathcal{A},\mathbf{Z}, \mathbf{R})$ as the city group graph, where $\mathcal{V}$ denotes the set of city group nodes, $\mathcal{A}$ denotes the set of edges, $\mathbf{Z}$ denotes the node attribute matrix, and $\mathbf{R}$ denotes the edge attribute matrix.

We construct the city graph during the pre-processing phase, while the city group graph will be constructed in subsequent processes. The construction rules of the city graph are as follows: 
\begin{align}
d_{i, j}=d_{j, i}=\sqrt{\left(x_{i}-x_{j}\right)^{2}+\left(y_{i}-y_{j}\right)^{2}} \\
\mathbf{E}_{i, j}=\frac{1}{d_{i, j}}, 0<d_{i, j}<R_{\text{h}}
\end{align}
where $[x_i,y_i]$ and $[x_j,y_j]$ denote the locations of cities $c_i$ and $c_j$, respectively. $d_{i,j}$ and $d_{j,i}$ denote the Euclidean distance between city $c_i$ and city $c_j$, and $R_\text{h}$ is the distance threshold. Only two cities with distance less than $R_\text{h}$ have connected on the city graph. $\mathbf{E}_{i,j}$ denotes the edge attributes of the edge from city $c_i$ to city $c_j$, and $\mathbf{E}_{i,j}$ and $\mathbf{E}_{j,i}$ are symmetrical.

\textbf{AQI data:} The air quality index (AQI) of city $c_i$ at time slot $t$ is represented as $aqi_i^t$.

\textbf{Weather data:} The weather data of city $c_i$ at time slot $t$ includes humidity, rainfall, air pressure, humidity, temperature, wind speed, and wind direction, denoted as $\boldsymbol{weather}_i^t$. Following \cite{p29}, we encode the wind direction data as a two-bit vector, and the encoding rules are shown in Table \ref{table_1}.

\begin{table}[!t]
\caption{Encoding Rules of Wind Directions}
\label{table_1}
\centering
\begin{tabular}{cc}
\hline
Direction &	Vector\\
\hline
North &	[0,1]\\
Northeast &	[1,1]\\
East &	[1,0]\\
Southeast &	[1,-1]\\
South &	[0,-1]\\
Southwest &	[-1,-1]\\
West &	[-1,0]\\
Northwest &	[-1,1]\\
No sustained direction &	[0,0]\\
\hline
\end{tabular}
\end{table}

\textbf{Historical observation data:} The historical observation data of city $c_i$ at time slot $t$ is denoted as $\boldsymbol{h}_i^t$, which consists of AQI data $aqi_i^t$ and weather data $\boldsymbol{weather}_i^t$. Thus the historical observation sequence of city $c_i$ at time slot $t$ is denoted as $H_i^{t-\tau_{\text{in}}+1}=\{\boldsymbol{h}_i^{t-\tau_{\text{in}}+1},\boldsymbol{h}_i^{t-\tau_{\text{in}}+2},\cdots,\boldsymbol{h}_i^t\}$, where $\tau_{\text{in}}$ is the historical window length.

\textbf{Time data:} The month, week, and day information of time slot $t$ are represented as one-hot encoding vectors, which are embedded end-to-end in the model as month vector $\boldsymbol{month}^t$, week vector $\boldsymbol{week}^t$, and hour vector $\boldsymbol{hour}^t$. The time vector of time slot $t$ is denoted as $\boldsymbol{time}^t$, concatenating the above three vectors.

\textbf{Nationwide city air quality forecasting:} Given the city locations $\mathbf{L}$, the historical observation sequence $H_i^{t-\tau_{\text{in}}+1:t}$, and the time vector $\boldsymbol{time}^t$ of time slot $t$, nationwide city air quality forecasting task aims to forecast the next $\tau_{\text{out}}$ AQI values for all cities, where $\tau_{\text{out}}$ denotes the forecasting horizon.

\subsection{Framework}

\begin{figure*}[htbp]
\begin{center}
\includegraphics{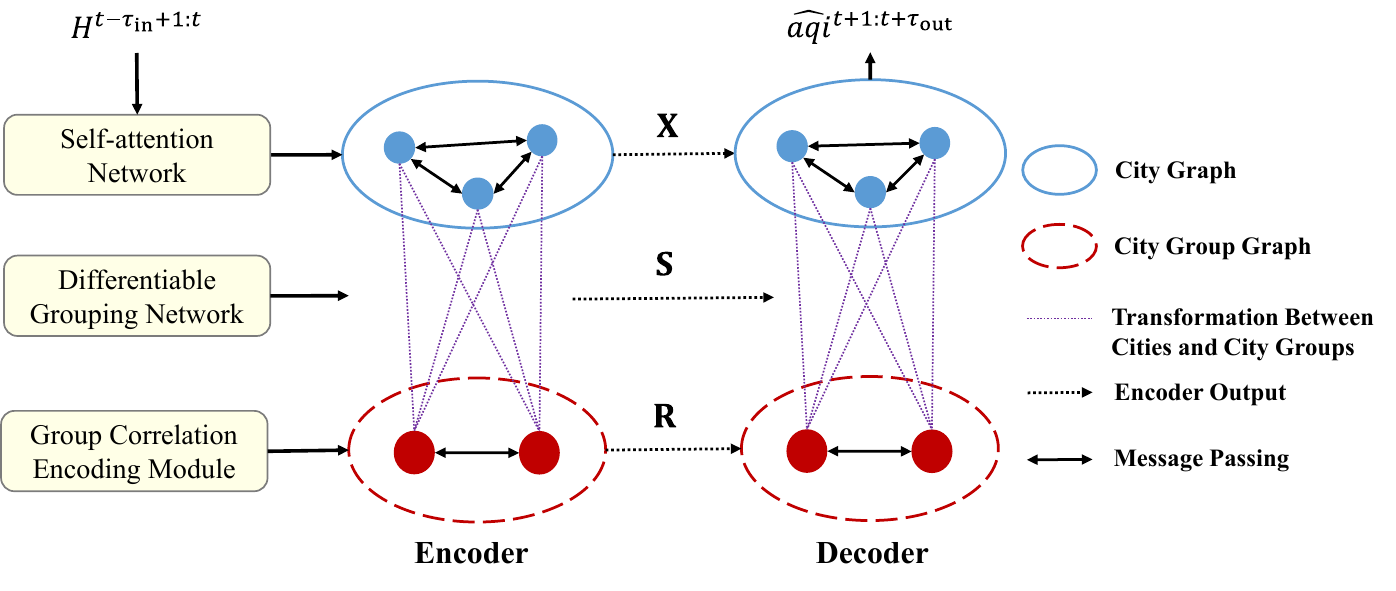}
\end{center}
\caption{Framework of GAGNN.}
\label{figure_1}
\end{figure*}

Figure \ref{figure_1} shows the framework of GAGNN, which leverages an encoder-decoder architecture. For the encoder, GAGNN utilizes self-attention network to extract the features of the historical observation sequence and obtain the city representations. In addition, GAGNN introduces differentiable grouping network to discover the latent dependencies among cities and groups cities to several city groups by a learning method. Based on the generated city groups, a group correlation encoding module is introduced to learn the correlations between them, which can better capture the dependencies between city groups. GAGNN implements the message passing mechanism in the city graph and the city group graph to model the dependencies between cities and city groups, respectively. The architecture of the decoder is similar to that of the encoder. The decoder forecasts the AQI values of all nationwide cities based on the outputs generated by the encoder, i.e., the updated city representations, the mapping relationships between cities and city groups, and the encoded correlations between city groups. 

\subsection{Sequence Feature Extraction}
\begin{figure}[htbp]
\begin{center}
\includegraphics{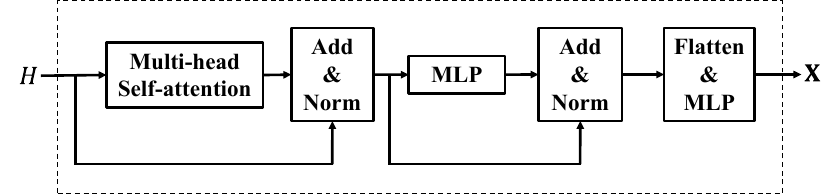}
\end{center}
\caption{Architecture of self-attention network.}
\label{figure_2}
\end{figure}

Self-attention network accepts the historical observation sequence $H^{t-\tau_{\text{in}}+1:t}$ as input and extracts features to obtain the city representations $\mathbf{X}$. As shown in Figure \ref{figure_2}, the design of self-attention network refers to the encoder architecture of Transformer \cite{p24}, where multi-head self-attention module implements the point-wise attention operation with multiple different sets of parameters. The computational process of the self-attention mechanism is defined as follows:
\begin{align}
\mathbf{Q}=f(H, \mathbf{W}_{\text {query}}),\mathbf{K}=f(H, \mathbf{W}_{\text {key}}),\mathbf{V}=f(H, \mathbf{W}_{\text {value}})
\end{align}
\begin{align}
{Attention}(\mathbf{Q}, \mathbf{K}, \mathbf{V})={softmax}\left(\frac{\mathbf{Q} \mathbf{K}^{\mathrm{T}}}{\sqrt{d_{\mathrm{key}}}}\right) \mathbf{V}
\end{align}
where $H$ denotes the historical observation sequence $H^{t-\tau_{\text{in}}+1:t}$, and transformed to query matrices $\mathbf{Q}$, key matrices $\mathbf{K}$, and value matrices $\mathbf{V}$. Here, $\mathbf{W}_{\text{query}}$, $\mathbf{W}_{\text{key}}$, $\mathbf{W}_{\text {value}}$ are learnable parameters and $d_\text{key}$ is the dimension of keys.

After the point-wise attention operation implemented by matrix multiplication, a sequence of vectors is obtained. After that, layer normalization and skip connection design are introduced, which can stabilize the output distribution of the network to reduce the difficulty of model training.

\subsection{City Grouping} \label{3_4}
Geographically distant cities may have strong correlations, but introducing multi-layer GNNs to model the latent dependencies between distant cities would cause the over-smoothing issue. Differentiable grouping network utilizes a learning method to capture the mapping relationships between cities and city groups, and generates city groups to discover the latent dependencies between cities. In this way, cities with strong latent dependencies would share the city group representations.

\begin{figure}[htbp]
\begin{center}
\includegraphics{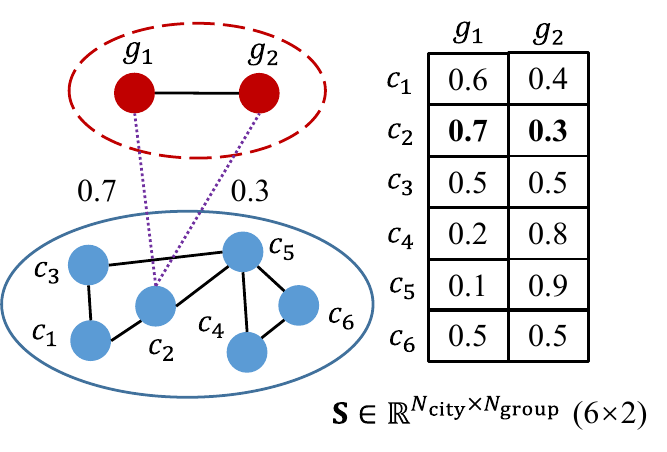}
\end{center}
\caption{Mapping relationships between cities and city groups in a case.}
\label{figure_3}
\end{figure}

Specifically, differentiable grouping network utilizes assignment matrix $\mathbf{S}\in \mathbb{R}^{N_{\text{city}} \times N_{\text{group}}}$ to indicate the mapping relationships between cities and city groups, where $\mathbf{S}_{i,j}$ denotes the probability of assigning $i$-th city to $j$-th city group. Thus, we have $\sum_{k=1}^{N_{\text{group}}}\mathbf{S}_{i,k} = 1$. $\mathbf{S}$ is randomly initialized and would be optimized during the training phase. A city can be assigned to multiple city groups with weights representing the relevance between the city and different city groups. Figure \ref{figure_3} further clarifies $\mathbf{S}$ with a case, in which there are 6 cities and 2 city groups. The probability of assigning city $c_2$ to city group $g_1$ is 0.7 and the probability of assigning city $c_2$ to city group $g_2$ is 0.3, which indicates that city $c_2$ is more correlated with city group $g_1$ than city group $g_2$.

The transformation from the city representations to the city group representations is achieved by $\mathbf{S}$. In addition, GAGNN introduces the geographic locations of cities in the transformation as their spatial features to capture the latent dependencies between adjacent cities, which is defined as follows:
\begin{align}
\mathbf{X}_{i}^{\prime}&=f_{\text{v}}\left(\mathbf{X}_{i}, \mathbf{L}_{i}\right) \\
\mathbf{Z}_{j}&=\sum_{i=1}^{N_{\text{city}}} \mathbf{s}_{j,i}^{\mathrm{T}} \mathbf{X}_{i}^{\prime}
\end{align}
where $\mathbf{X}_{i}$ is the output of self-attention network, $\mathbf{L}_{i}$ is the geographic location of city $c_i$, $\mathbf{X}_{i}^{\prime}$ is the city representation of city $c_i$ containing geographic information, $\mathbf{Z}_{j}$ is the city group representation of city group $g_j$ obtained by cities assigned to it, $\mathbf{S}^T$ denotes the transpose matrix of $\mathbf{S}$, and $f_{\text{v}}$ is a fusion function implemented by a multi-layer perceptron (MLP).

\subsection{Modeling the Dependencies Between City Groups}
GAGNN introduces a group correlation encoding module to complete the construction of the city group graph. After that, GAGNN implements message passing mechanism in the city group graph to model the dependencies between city groups.

Existing models \cite{p14,p30,p33} defines the correlations between city groups by the correlations between the cities assigned to them, e.g., the geographic distance, which cannot effectively capture the dependencies between city groups.

To address this problem, GAGNN introduces a group correlation encoding module, which encodes the edge attributes between city group nodes by a learning method. Specifically, the group correlation encoding module considers not only the city group representations, but also other factors affecting the correlations between city groups, e.g., time information. We construct the city group graph as a fully connected graph. Considering that $N_\text{group}$ is much smaller than $N_\text{city}$, the computational cost to encode the correlations between pair-wise city groups is acceptable. The process of group correlation encoding is defined as follows:
\begin{equation}
\mathbf{R}_{i, j}={ReLU}\left({enc}\left(\mathbf{Z}_{i}, \mathbf{Z}_{j}, \boldsymbol{time}\right)\right)
\end{equation}
where $\mathbf{R}_{i, j}$ denotes the edge attributes of the edge between city groups $g_i$, $g_j$, and $\mathbf{Z}_{i}$, $\mathbf{Z}_{j}$ are their representations, respectively. $\boldsymbol {time}$ is the time vector, which affects the correlations between city groups, and ${enc}$ is a encoding function implemented by a MLP.

Message passing mechanism is implemented to model the dependencies between city groups, which consists of two major processes: message aggregation and representation updating. The details are shown as follows:
\begin{align}
R_{i}&=\left\{\left(\mathbf{Z}_{i}, \mathbf{Z}_{j}, \mathbf{R}_{j, i}\right)\right\}_{i \neq j} \\
\boldsymbol{r}_{i} &\leftarrow \rho_{\text{g}} \left(R_{i}\right) \\
\mathbf{Z}_{i}^{\prime} &\leftarrow \phi_{\text{g}}\left(\boldsymbol{r}_{i}, \mathbf{Z}_{i}\right)
\end{align}
where $R_i$ is the set containing all the messages passed to city group $g_i$, which would be transformed to vector $\boldsymbol{r}_{i}$ later, $\mathbf{Z}_i^\prime$ is the updated city group representation of city group $g_i$ based on $\boldsymbol{r}_{i}$, and $\rho_{\text{g}}$, $\phi_{\text{g}}$ are transformation functions implemented by MLPs.

\subsection{Modeling the Dependencies Between Cities}
To model both the spatial and latent dependencies between cities, GAGNN updates the representations of cities based on the city groups they assigned to and implements the message passing mechanism in the city graph.

Similar to the transformation process in Section \ref{3_4}, the group based representations of cities can be obtained as follows:
\begin{align}
\mathbf{X}_{i}^{1}=\sum_{j=1}^{N_{\text {group}}} \mathbf{S}_{i, j} \mathbf{Z}_{j}^{\prime}
\end{align}
where $\mathbf{Z}_{j}^{\prime}$ is the updated representation of city group $g_j$ and $\mathbf{X}_{i}^{1}$ is the group based representation of city $c_i$.

Slightly different from the calculation process in the city group graph, the message passing mechanism in the city graph firstly fuses the group based representation of city and the representation obtained from self-attention network. In this way, the fused city representations contain the information of local features and the city groups assigned to.

The message aggregation and representation updating processes are similar to those in the city group graph. The details are shown as follows:
\begin{align}
\mathbf{X}_{i}^{2}&={cat}\left(\mathbf{X}_{i}, \mathbf{X}_{i}^{1}\right) \\
R_{i}&=\left\{\left(\mathbf{X}_{i}^{2}, \mathbf{X}_{n}^{2}, \mathbf{E}_{n, i}\right)\right\}_{n \in \mathcal{N}(i)} \\
\boldsymbol{r}_{i} &\leftarrow \rho\left(R_{i}\right) \\
\mathbf{X}_{i}^{3} &\leftarrow \phi\left(\mathbf{X}_{i}^{2}, \boldsymbol{r}_{i}\right)
\end{align}
where $\mathbf{X}_{i}^{2}$ is the fused representations of city $c_i$ containing the information of local features and city groups, $R_i$ is the set containing all the messages passed to city $c_i$ from its neighbors, which would be transformed to vector $\boldsymbol{r}_{i}$ later, and $\mathbf{X}_{i}^{3}$ is the updated city representation of city $c_i$ based on $\boldsymbol{r}_{i}$. Concatenation function $cat$ and transformation functions $\rho$, $\phi$ are implemented by MLPs.

\subsection{Forecasting and Learning}
GAGNN leverages an encoder-decoder architecture, and generates AQI forecasting results in the decoder.

The decoder accepts the outputs generated by the encoder, i.e., $\mathbf{X}^{3}$, $\mathbf{S}$ ($\mathbf{S}$ in the decoder does not require gradient), and $\mathbf{R}$ (the correlations would not be re-encoded). The calculation steps in the decoder are similar to the encoder, with the difference that the inputs to the decoder are $\mathbf{X}^{3}$ rather than the historical observation sequence, thus self-attention network is omitted.

After the calculation process in the decoder, we get the final city representations $\mathbf{X}^{\text {output }}$, and further forecast the AQI values of all cities:
\begin{equation}
\widehat{aqi}_{i}^{t+1: t+\tau_{\mathrm{out}}}={for}\left(\mathbf{X}_{i}^{\text {output }}\right)
\end{equation}
where ${for}$ is a forecasting function implemented by a MLP.

We using mean absolute error (MAE) to evaluate the error between true AQI values and forecasted AQI values, and the loss function is defined as follows:
\begin{equation}
L(\theta)=\frac{1}{N_{\text {city }} \times \tau_{\text {out }}} \sum_{i=1}^{N_{\text {city }}} \sum_{k=1}^{\tau_{\text {out }}}\left|a q i_{i}^{t+k}-\widehat{a q i}_{i}^{t+k}\right|
\end{equation}
where $\widehat{aqi}_{i}^{t+k}$ denotes the forecasted AQI value of city $c_i$ at time slot $t+k$, ${aqi}_{i}^{t+k}$ denotes the true AQI value of city $c_i$ at time slot $t+k$, and $\theta$ denotes the learnable parameters in GAGNN.

\section{Experiments}
In this section, we evaluate the performance of GAGNN on Chinese city air quality dataset, which covers 209 cities over a period of 850 days. The details of the dataset and experimental settings are provided in Section \ref{4_0} and Section \ref{4_0_2}. We investigate how the number of city groups $N_{\text{group}}$ affects the performance of GAGNN in Section \ref{4_1}. The effectiveness of model components are studied in Section \ref{4_2}. We compare GAGNN with other forecasting models in Section \ref{4_3}, and provide a case to illustrate the superior performance of GAGNN in Section \ref{4_4}.

Besides MAE, we also used root mean squared error (RMSE) to evaluate the model performance, which is more sensitive to the deviations:
\begin{equation}
R M S E=\sqrt{\frac{1}{N_{\text {city }}} \sum_{i=1}^{N_{\text {city}}}\left|a q i_{i}^{t+k}-\widehat{a q i}_{i}^{t+k}\right|^{2}}
\end{equation}

In the experiments, we utilize the previous 24-hour observations of all cities to forecast the next 6-hour AQI values, i.e., $\tau_{\text{in}}=24$ and $\tau_{\text{out}}=6$. All experiments were repeated five times under the same settings to avoid contingency and take the average of five experimental results to evaluate the performance.

\subsection{Datasets} \label{4_0}
We evaluate the performance of GAGNN on Chinese city air quality dataset. The dataset contains the AQI data and weather data of 209 cities, which are collected from January 1, 2017 to April 30, 2019. The details of these data are described as follows:

(1) AQI data, including AQI values and the locations of cities, are collected from National Urban Air Quality Real-time Release Platform\footnote{http://106.37.208.233:20035/}. We collect AQI values at 1-hour granularity.

(2) Weather data are collected from Envicloud\footnote{http://www.envicloud.cn/}, a data service provider. We collect weather data at 1-hour granularity.

\begin{figure}[htbp]
\begin{center}
\includegraphics{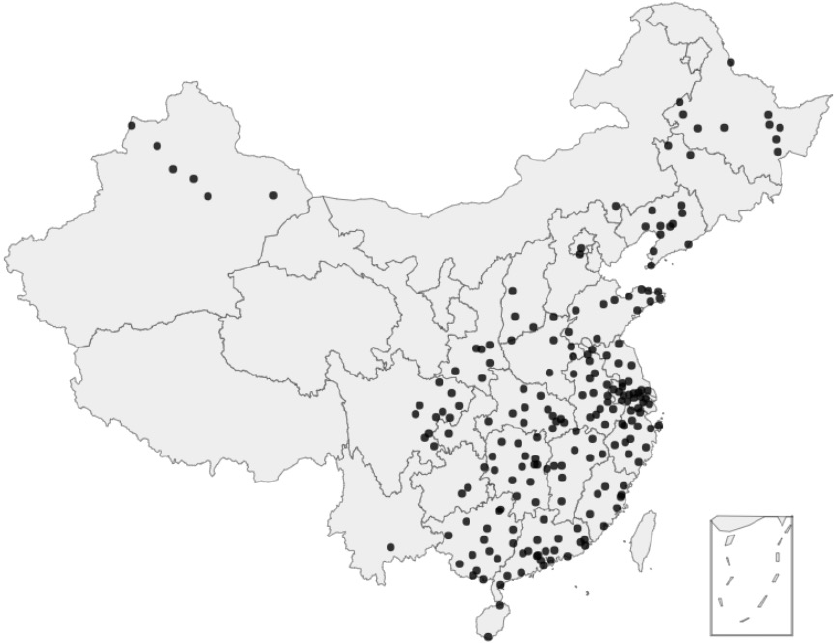}
\end{center}
\caption{Geographical locations of 209 cities.}
\label{figure_7}
\end{figure}

The geographical locations of all cities are shown in Figure \ref{figure_7}, identified by black dots on the map. Cities with significant AQI or weather data missing are not included in the dataset. The few remaining missing values in the dataset are filled in by linear interpolation method. Sliding windows (step = 1 hour) are used to generate samples, and we finally get 20,370 samples of 209 cities.

\subsection{Experimental Settings} \label{4_0_2}
We chronologically split all the samples into training data, validation data, and test data by the radio of 0.7:0.1:0.2. Through the hyperparameter evaluation (Section \ref{4_1}), the number of city groups $N_{\text{group}}$ is set to 15. In addition, the batch size is set to 64, the epoch number is set to 300, the hidden size of GNNs is set to 32, the layer number of GNNs is set to 2, the edge attribute dimension of the city group graph is set to 12, and the distance threshold $R_\text{h}$ is set to 250 KM. We choose Adam \cite{p9} as the optimization method. The learn rates for the parameters of $\mathbf{S}$ and other parameters are 0.05 and 0.001, respectively.

We implement our model with PyTorch \cite{p19} and construct GNNs with PyTorch Geometric library \cite{p5}. The sourse code is released on GitHub\footnote{https://github.com/Friger/GAGNN}. We implement training, validation, and test tasks on a server with one GPU (NVIDIA RTX 2080Ti).

\subsection{Hyperparameter Evaluation} \label{4_1}
The number of city groups $N_{\text{group}}$ is a hyperparameter that needs to set in advance. Based on the process introduced in Section \ref{3_4}, differentiable grouping network would assign all cities to $N_{\text{group}}$ city groups.

\begin{figure}[htbp]
\begin{center}
\includegraphics{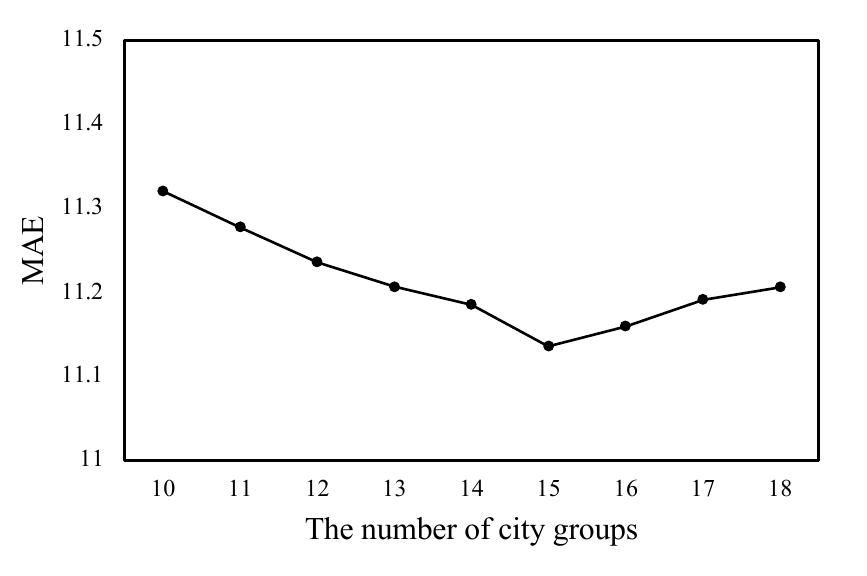}
\end{center}
\caption{Evaluation results of different $N_{\text{group}}$.}
\label{figure_4}
\end{figure}

We evaluate the effects of $N_{\text{group}}$ with the average MAE on validation data. Figure \ref{figure_4} shows the evaluation results of different $N_{\text{group}}$ values varying from 10 to 18.

The results indicate that GAGNN performs optimally when $N_{\text{group}}$ equals to 15. As $N_{\text{group}}$ increases, MAE decreases first and then increases. When $N_{\text{group}}$ is too small, the dependencies among cities cannot be fully exploited, while when $N_{\text{group}}$ is too large, the distribution of city grouping results will be more dispersed, increasing the difficulty of model training.

\begin{figure*}[htbp]
\begin{center}
\includegraphics{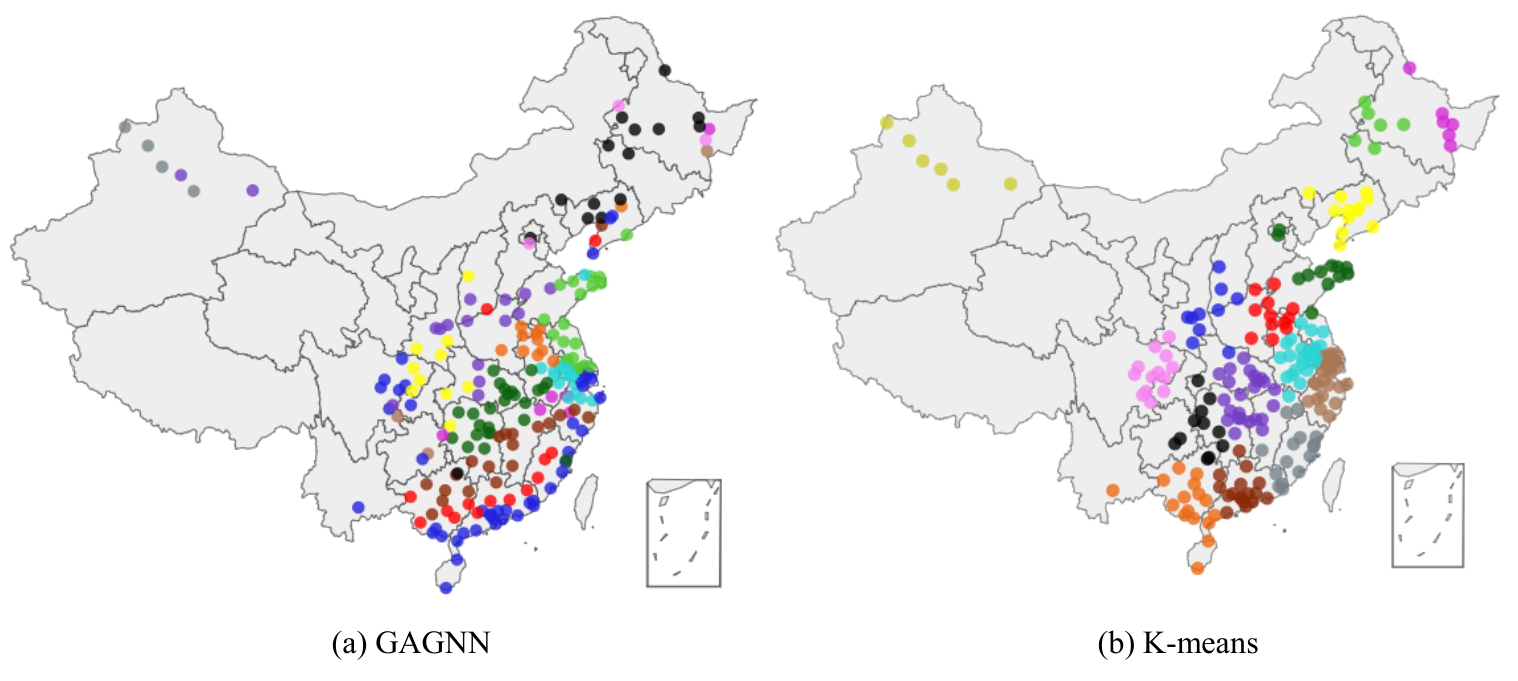}
\end{center}
\caption{Results of city grouping.}
\label{figure_5}
\end{figure*}

K-means algorithm can be considered as a graph pooling method based on the geographical distribution of cities. In Figure \ref{figure_5}, we give the grouping visualization ($N_{\text{group}}=15$ for GAGNN and $k=15$ for K-means), where different city groups are distinguished by different colors, and there is no correspondence between the city groups with the same color. Specifically, we mark cities by the city group that has the highest probability in $\mathbf{S}$ for GAGNN.

It can be seen that differentiable grouping network can discover some latent dependencies among cities that K-means algorithm is unable to detect. For example, due to the sea breeze effect \cite{p18}, cities distributed in a strip distribution along the southern coast have strong correlations, while K-means algorithm cannot capture this effect, and southern coastal cities are assigned to different city groups.

\subsection{Model Component Evaluation} \label{4_2}
To verify the effectiveness of the components introduced in GAGNN, one of these components was removed or modified at a time in model component evaluation. To ensure fairness, all variant models follow the same experimental settings as GAGNN. We compare GAGNN with the following variants:

GAGNN with LSTM (G with LSTM): G with LSTM replaces self-attention network with a LSTM to extracts the features of the historical observation sequence.

GAGNN with K-means (G with K-means): G with K-means uses K-means algorithm instead of a learning method to obtain the mapping relationships between cities and city groups.

GAGNN without the group correlation encoding module (G w/o CE): G w/o CE removes the group correlation encoding module in the city group graph. We construct the city group graph as a fully connected graph without edge attributes.

GAGNN without location (G w/o loc): G w/o loc removes the spatial features, i.e., the geographic locations of cities, introduced in the calculation of city group representations.

\begin{table}[!t]
\caption{Results of Model Component Evaluation}
\label{table_2}
\centering
\resizebox{0.48\textwidth}{!}{
\begin{tabular}{llllllll}
\hline
Model                           & Metric & 1h    & 2h    & 3h    & 4h    & 5h    & 6h    \\ \hline
\multirow{2}{*}{GAGNN}          & MAE    & \textbf{5.56}  & \textbf{8.59 } & \textbf{10.80 }& \textbf{12.52} & \textbf{13.91} & \textbf{15.10} \\
                                & RMSE   & \textbf{10.81} & \textbf{16.17} & \textbf{19.84} & \textbf{22.51} & \textbf{24.63} & \textbf{26.37} \\ \cline{2-8}
\multirow{2}{*}{G with LSTM}    & MAE    & 5.86  & 9.01  & 11.34 & 13.12 & 14.57 & 15.81 \\
                                & RMSE   & 11.30 & 16.83 & 20.74 & 23.53 & 25.70 & 27.43 \\\cline{2-8}
\multirow{2}{*}{G with K-means} & MAE    & 5.91  & 9.15  & 11.53 & 13.55 & 15.02 & 16.10 \\
                                & RMSE   & 11.67 & 17.15 & 20.94 & 23.94 & 26.36 & 27.94 \\\cline{2-8}
\multirow{2}{*}{G w/o CE}       & MAE    & 5.74  & 8.78  & 11.05 & 12.84 & 14.29 & 15.54 \\
                                & RMSE   & 11.16 & 16.52 & 20.29 & 23.07 & 25.29 & 27.13 \\\cline{2-8}
\multirow{2}{*}{G w/o loc}      & MAE    & 5.69  & 8.72  & 10.96 & 12.70 & 14.12 & 15.37 \\
                                & RMSE   & 11.05 & 16.35 & 20.06 & 22.79 & 24.98 & 27.10 \\ \cline{1-8} 
\end{tabular}}
\end{table}

The performances of GAGNN and its variants are given in Table \ref{table_2}, and the following tendencies can be discovered:

(1) GAGNN outperforms G with LSTM on all metrics. The result indicates that self-attention network is a better practice to extract sequence features, as self-attention network can model the point-wise correlations of elements in the sequence.

(2) GAGNN outperforms G with K-means on all metrics. The result indicates that the grouping results obtained by K-means cannot fully discover the latent dependencies among cities. GAGNN introduces a learning method to group cities to discover the latent dependencies among cities.

(3) GAGNN outperforms G w/o CE on all metrics. The result indicates that the group correlation encoding module introduced in GAGNN can better capture the dependencies between city groups.

(4) GAGNN outperforms G w/o loc on all metrics. The result indicates that introducing the geographic locations of cities as spatial features in city grouping can capture the latent dependencies between adjacent cities.

\subsection{Comparison with Other Forecasting Models} \label{4_3}
To further verify the effectiveness of our model, we compare GAGNN with existing forecasting models. We selected the following types of models for comparison: classical regression models (LSTM, XGBoost), flatten structure forecasting models (FGA, GC-LSTM), an architecture enhanced GNN model (DeeperGCN), and hierarchical structure forecasting models (ST-UNet, SHARE).

\textbf{Classical regression models:}

LSTM: LSTM \cite{p8} introduces the gating mechanism based on RNN to relieve the gradient problem. We use the historical observation sequences as the LSTM input and forecast AQI values. LSTMs for all cities share the parameters. After optimization, the hidden unit of LSTM is set to 32.

XGBoost: XGBoost \cite{p3} is an engineered implementation of GBRT, which introduces some strategies to support parallel computation and outlier handling. A separate XGBoost model is built to forecast each time slot for each city. We use grid search method to optimize the hyperparameters.

\textbf{Flatten structure forecasting models:}

Flatten GAGNN (FGA): FGA removes the hierarchical structure in GAGNN. Other experiment settings follow the original model.

GC-LSTM: GC-LSTM \cite{p20} constructs a flatten city graph based on the geographical distribution of cities and introduces GCN and LSTM to capture spatial and temporal dependencies, respectively. After optimization, the hidden unit of LSTM is set to 64, and the output dimension of GCN is set to 32.

\textbf{Architecture enhanced GNN model:}

DeeperGCN: DeeperGCN \cite{p13} introduces a pre-activation architecture and a message normalization method to relieve the over-smoothing issue when GNNs go deeper. After optimization, the layer number of DeeperGCN is set to 8.

\textbf{Hierarchical structure forecasting models:}

ST-UNet: ST-UNet \cite{p31} adopts a heuristic method to construct coarsened graphs and adopts pooling and unpooling strategies to implement inter-level interactions. In addition, ST-UNet introduces dilated GRUs to capture the multilevel temporal dependencies of sequences. After optimization, the hidden unit of GRU is set to 32, and the output dimension of GCN is set to 32.

SHARE: SHARE \cite{p33} is a semi-supervised spatial temporal forecasting model, which introduces a soft pooling method to construct a coarsened graph and concatenates the entity representations in multi-level graphs to obtain forecast results. The correlations between coarsened nodes are defined by the correlations between the nodes assigned to them. We remove the semi-supervised learning part of SHARE and remain the rest settings.

The performances of GAGNN and other forecasting models are given in Table \ref{table_3}, and the following tendencies can be discovered:

\begin{table}[!t]
\caption{Results of Different Forecasting Models}
\label{table_3}
\centering
\resizebox{0.48\textwidth}{!}{
\begin{tabular}{llllllll}
\hline
Model                      & Metric & 1h    & 2h    & 3h    & 4h    & 5h    & 6h    \\\hline
\multirow{2}{*}{GAGNN}     & MAE    & \textbf{5.56}  & \textbf{8.59}  & \textbf{10.80} & \textbf{12.52} & \textbf{13.91} & \textbf{15.10} \\
                           & RMSE   & \textbf{10.81} & \textbf{16.17} & \textbf{19.84} & \textbf{22.51} & \textbf{24.63} & \textbf{26.37} \\  \cline{2-8}
\multirow{2}{*}{LSTM}      & MAE    & 6.50  & 10.26 & 13.18 & 15.52 & 17.40 & 18.91 \\
                           & RMSE   & 13.85 & 19.26 & 23.52 & 26.83 & 29.46 & 31.55 \\ \cline{2-8}
\multirow{2}{*}{XGBoost}   & MAE    & 6.85  & 10.89 & 13.99 & 16.27 & 18.14 & 19.56 \\
                           & RMSE   & 14.25 & 19.80 & 24.72 & 28.14 & 30.63 & 33.44 \\ \cline{2-8}
\multirow{2}{*}{FGA}       & MAE    & 5.87  & 9.14  & 11.71 & 13.75 & 15.42 & 16.80 \\
                           & RMSE   & 11.36 & 17.01 & 21.05 & 24.09 & 26.55 & 28.52 \\ \cline{2-8}
\multirow{2}{*}{GC-LSTM}   & MAE    & 5.95  & 9.16  & 11.58 & 13.46 & 15.00 & 16.31 \\
                           & RMSE   & 11.91 & 16.98 & 20.82 & 23.69 & 25.97 & 27.82 \\ \cline{2-8}
\multirow{2}{*}{ST-UNet}   & MAE    & 5.95  & 9.30  & 11.58 & 13.38 & 14.82 & 16.02 \\
                           & RMSE   & 11.74 & 18.01 & 21.34 & 23.90 & 25.94 & 27.64 \\ \cline{2-8}
\multirow{2}{*}{SHARE}     & MAE    & 5.84  & 9.07  & 11.49 & 13.35 & 14.74 & 15.79 \\
                           & RMSE   & 11.27 & 16.84 & 20.77 & 23.60 & 25.80 & 27.38 \\ \cline{2-8}
\multirow{2}{*}{DeeperGCN} & MAE    & 6.54  & 9.74  & 11.77 & 13.40 & 15.29 & 16.41 \\
                           & RMSE   & 13.67 & 18.93 & 21.14 & 23.83 & 26.25 & 28.02 \\\hline
\end{tabular}}
\end{table}

(1) Flatten structure forecasting models outperform classical regression models. Classical regression models tend to model the relationship between single city features and air quality, ignoring the dependencies among cities. To address this problem, flatten structure forecasting models introduce GNNs to capture the dependencies among cities.

(2) DeeperGCN outperforms flatten structure forecasting models. It indicates that modeling the distant cities with strong correlations can improve forecasting accuracy. DeeperGCN improves the architectural designs of GNNs to expand the receptive field of cities without raising the over-smoothing issue.

(3) Hierarchical structure forecasting models outperform DeeperGCN. It indicates that the hierarchical structure forecasting models can relieve the conflict between expanding receptive fields and preserving local features, while the receptive field size is not adaptive for each entity in DeeperGCN, which cannot effectively capture the dependencies between entities.

(4) SHARE outperforms ST-UNet. The major difference between SHARE and ST-UNet is that SHARE utilizes a soft pooling method to obtain the mapping relationships between cities and city groups, while ST-UNet uses a heuristic grouping method based on the geographical distribution of cities. The results indicate that a learning grouping method can effectively discover the latent dependencies between cities.

(5) GAGNN outperforms SHARE. The major difference between GAGNN and SHARE is that GAGNN introduces a group correlation encoding module to learn the correlations between city groups, while in SHARE they are defined by the geographical distribution of cities assigned to. The group correlation encoding module encodes the edge attributes between city group nodes based on the city group representations and time information, which can effectively capture the dependencies between city groups.

\subsection{Case Study} \label{4_4}
\begin{figure}[htbp]
\begin{center}
\includegraphics{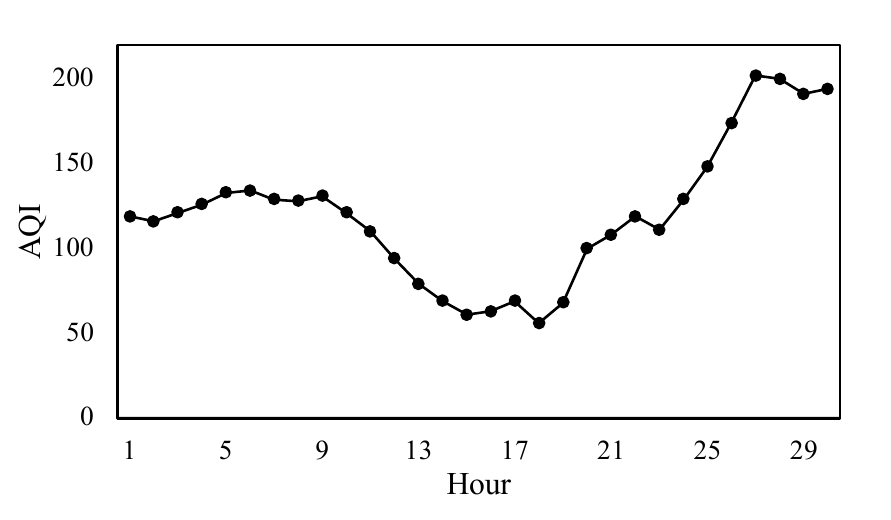}
\end{center}
\caption{AQI values of Beijing city in the case.}
\label{figure_6}
\end{figure}

In case study, we give an example to illustrate the superiority of GAGNN over existing models when regional pollution occurs and the correlations among cities are complex. The task is to forecast the AQI values of Beijing city from 18:00 November 24 to 23:00 November 24, 2018. Figure \ref{figure_6} shows the AQI values of Beijing city in the historical window (1-24 hours) and the forecasting horizon (25-30 hours). In this case, an air pollution event occurred in Beijing city and the AQI values in Beijing city increased sharply from 18:00 November 24, making the task of forecasting difficult.

Table \ref{table_4} gives the forecasting results of GAGNN and other models in this case. We can find that the forecasting errors of all models become larger than usual, as the dependencies in spatial and temporal dimensions are complicated when regional pollution occurs. Compared with other models, GAGNN achieves high accuracy in forecasting, which indicates that differentiable grouping network and the group correlation encoding module introduced in GAGNN can effectively capture the complex dependencies among cities.

\begin{table}[!t]
\caption{MAEs of Different Models in the Case}
\label{table_4}
\centering
\resizebox{0.48\textwidth}{!}{
\begin{tabular}{lllllll}
\hline
Model     & 1h    & 2h    & 3h    & 4h    & 5h    & 6h    \\\hline
GAGNN     & \textbf{1.79}  & \textbf{15.95} &\textbf{ 35.06 }& \textbf{37.00 }& \textbf{29.08} & \textbf{27.18} \\
LSTM      & 23.70 & 46.51 & 62.77 & 64.31 & 48.99 & 49.06 \\
XGBoost   & 21.91 & 33.16 & 60.81 & 60.44 & 47.20 & 50.06 \\
FGA       & 13.78 & 26.59 & 55.40 & 55.72 & 48.88 & 48.82 \\
GC-LSTM   & 11.22 & 27.61 & 50.09 & 58.90 & 45.32 & 46.65 \\
ST-UNet   & 8.66  & 24.45 & 48.34 & 49.20 & 45.97 & 47.26 \\
SHARE     & 10.23 & 23.32 & 48.93 & 50.78 & 49.68 & 44.59 \\
DeeperGCN & 14.19 & 24.04 & 49.91 & 55.63 & 40.79 & 45.82 \\\hline
\end{tabular}}
\end{table}

\section{Conclusions and future work}
We propose GAGNN, a hierarchical model for nationwide city air quality forecasting. The model constructs a city graph and a city-group graph to model the spatial and latent dependencies between cities, respectively. We evaluate GAGNN on Chinese city air quality dataset and obtain the following conclusions from the experiment results: (1) HGNNs can effectively relieve the conflict between expanding receptive fields and preserving local features; (2) differentiable grouping network can effectively discover the latent dependencies between cities; (3) the group correlation encoding module can effectively capture the dependencies between city groups.

In the future, we will extend our model in the following aspects. On the one hand, the current model considers only two levels of hierarchy, i.e., cities and city groups. We will introduce a multi-level hierarchical structure in the model to capture more complex correlations between entities. On the other hand, the current correlations between city groups obtained by a learning method do not have clear semantic information. We will introduce some constraints that allow the model to consider more semantic correlations between entities, e.g., causality.


\ifCLASSOPTIONcaptionsoff
  \newpage
\fi

\bibliographystyle{IEEEtran}
\bibliography{bare_jrnl.bib}
%
\begin{IEEEbiography}[{\includegraphics[width=1in,height=1.25in,clip,keepaspectratio]{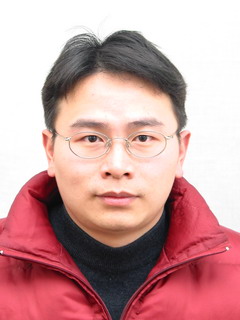}}]
{Ling Chen} received the B.S. and Ph.D. degrees in computer science from Zhejiang University, China, in 1999 and 2004, respectively. He is currently a Professor with the College of Computer Science and Technology, Zhejiang University, China. His research interests include ubiquitous computing and data mining.
\end{IEEEbiography}

\begin{IEEEbiography}[{\includegraphics[width=1in,height=1.25in,clip,keepaspectratio]{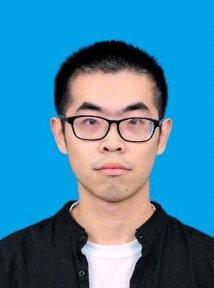}}]{Jiahui Xu}
received the B.Eng. degree in Communication Engineering from Wuhan University Of Technology, China, in 2018. He is currently a M.S. candidate with the College of Computer Science and Technology, Zhejiang University, China. His research interests include urban computing and time series modeling.
\end{IEEEbiography}

\begin{IEEEbiography}[{\includegraphics[width=1in,height=1.25in,clip,keepaspectratio]{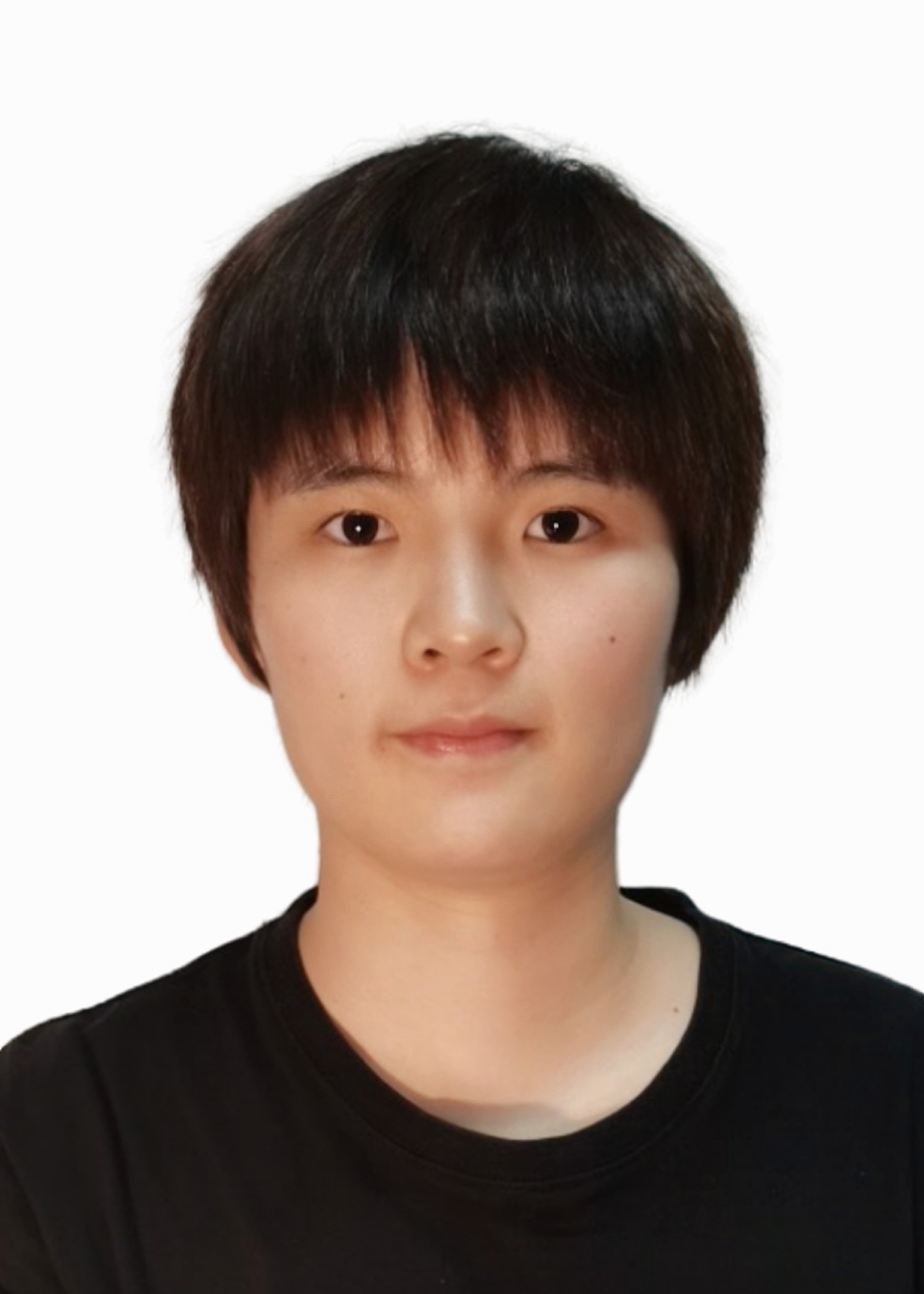}}]
{Binqing Wu} received the B.Eng. degree in computer science from Southwest Jiangtong University, China, in 2020. She is currently a Ph.D. student with the College of Computer Science and Technology, Zhejiang University, China. Her research interests include urban computing and data mining.
\end{IEEEbiography}

\begin{IEEEbiography}[{\includegraphics[width=1in,height=1.25in,clip,keepaspectratio]{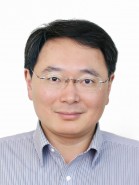}}]
{Yuntao Qian} (Member, IEEE) received the B.E. and M.E. degrees in automatic control from Xi’an Jiaotong University, China, in 1989 and 1992, respectively, and the Ph.D. degree in signal processing from Xidian University, in 1996. He is currently a Professor with the College of Computer Science and Technology, Zhejiang University, China. His research interests include machine learning, signal and image processing, pattern recognition, and hyperspectral imaging.

Dr. Qian is an Associate Editor of the IEEE JOURNAL OF SELECTED TOPICS IN APPLIED EARTH OBSERVATIONS AND REMOTE SENSING.
\end{IEEEbiography}

\begin{IEEEbiography}[{\includegraphics[width=1in,height=1.25in,clip,keepaspectratio]{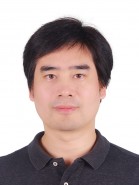}}]
{Zhenhong Du} (Member, IEEE) received the Ph.D. degree in cartography and geographic information science from Zhejiang University, China, in 2010. He is currently a Professor with the School of Earth Sciences, Zhejiang University, China. He is also the Vice-Dean of the School of Earth Sciences and the Director of the Institute of Geography and Spatial Information, Zhejiang University, China. His research interests include remote sensing and geographic information science, spatial–temporal big data and artificial intelligence, and big data and earth systems.
\end{IEEEbiography}

\begin{IEEEbiography}[{\includegraphics[width=1in,height=1.25in,clip,keepaspectratio]{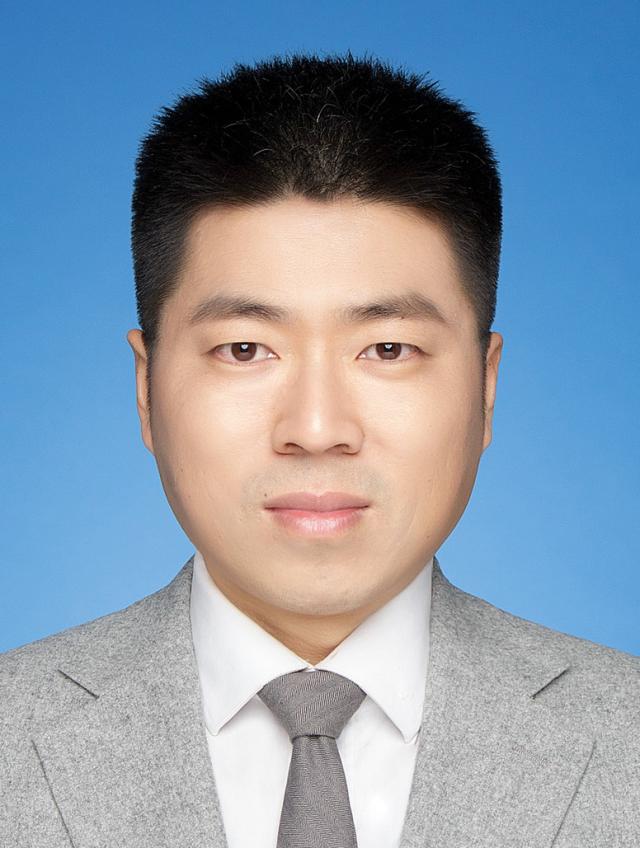}}]
{Yansheng Li} received the B.S. degree from the School of Mathematics and Statistics, Shandong University, China, in 2010, and the Ph.D. degree from the School of Automation, Huazhong University of Science and Technology, China, in 2015. He is currently an Associate Professor with the School of Remote Sensing and Information Engineering, Wuhan University, China. His research interests include computer vision, machine learning, deep learning, and their applications in remote sensing.
\end{IEEEbiography}

\begin{IEEEbiography}[{\includegraphics[width=1in,height=1.25in,clip,keepaspectratio]{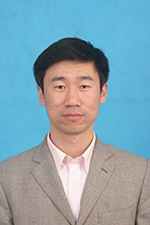}}]
{Yongjun Zhang} received the B.S., M.S., and Ph.D. degrees from Wuhan University, China, in 1997, 2000, and 2002, respectively. He is currently a Professor of photogrammetry and remote sensing with the School of Remote Sensing and Information Engineering, Wuhan University, China. His research interests include space, aerial, and lowattitude photogrammetry, image matching, combined bundle adjustment with multisource data sets, and 3-D city reconstruction.
\end{IEEEbiography}

\newpage

\end{document}